\pdfoutput=1

\documentclass[11pt]{article}

\usepackage[final]{coling}

\usepackage{times}
\usepackage{latexsym}

\usepackage[T1]{fontenc}

\usepackage[utf8]{inputenc}

\usepackage{microtype}

\usepackage{inconsolata}

\usepackage{graphicx}
\usepackage{booktabs}
\usepackage{multirow}
\usepackage{float}

\title{Not All Jokes Land: Evaluating Large Language Models’ Understanding of Workplace Humor}

\author{Mohammadamin Shafiei\thanks{\enspace Equal contribution.}\textsuperscript{1}, \space 
  Hamidreza Saffari$^{*}$\textsuperscript{2} \\
\textsuperscript{1}University of Milan, 
\textsuperscript{2}Politecnico di Milano \\
\texttt{m.shafieiapoorvari@studenti.unimi.it} \\
\texttt{hamidreza.saffari@mail.polimi.it} \\
}

\begin{document}
\maketitle

\begin{abstract}
With the recent advances in Artificial Intelligence (AI) and Large Language Models (LLMs), the automation of daily tasks, like automatic writing, is getting more and more attention. Hence, efforts have focused on aligning LLMs with human values, yet humor—particularly professional industrial humor used in workplaces—has been largely neglected. To address this, we develop a dataset of professional humor statements along with features that determine the appropriateness of each statement. Our evaluation of five LLMs shows that LLMs often struggle to judge the appropriateness of humor accurately. \citet{etxaniz2024bertaqa}
\end{abstract}

\section{Introduction}
Humor is a universal fundamental aspect of human expression, \citep{bardon2005philosophy} encompassing a broad range of emotions and interactions. \citep{morreall1983humor} From a societal point of view, humor has long served as a means of communication in society, while also being used to cope with life's challenges and create a sense of trust. \citep{meyer2000humor, meyer2015understanding, lynch2002humorous, berk2015greatest} Humor is often seen as spontaneous, yet it is strongly shaped by context, including cultural and societal backgrounds and the relationship between speaker and audience.\citep{jiang2019cultural, hampes1992relation}

Many studies have focused on the relationship between the setting and humor. \citep{vizmuller1980psychological, banas2011review, coser1959some, zhou-etal-2023-cobra} In professional industrial environments, humor takes on unique dynamics, reflecting organizational structures and interpersonal relationships norms. \citep{mesmer2012meta, idrees2020styles} When used well, it strengthens teams and fosters innovation; misused, risks misunderstandings and credibility. \citep{meyer2000humor}
\begin{figure}[t]
  \includegraphics[width=\columnwidth]{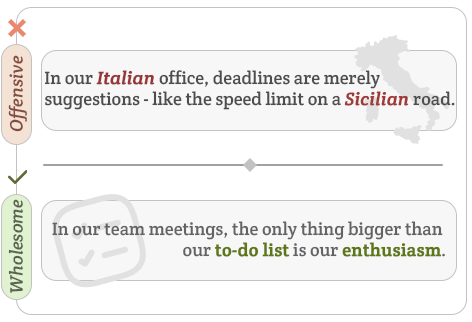}
  \caption{Examples of humor misclassified by LLMs.}
  \label{fig:examples}
\end{figure}
Also, from the computational side, with the advances in Artificial intelligence (AI) and more specifically Large Language Models (LLMs), many efforts have been dedicated to automating various tasks like writing and content creation.\citep{wang2024investigating, xiao2024automation, scherbakov2024emergence, shen2024data} An example of automated writing is the use of LLM agents to write an email, a document, or a social media post, all applicable to industrial settings. \citep{jovic2024evaluating, gryka2024detection, lu2024corporate} Another area that has also got attention to make automation is automating simulation of human life and interactions using persona-assigned LLM agents. \citep{park2023generative, wang2024survey} In these automation scenarios, LLMs may use humor and need situational awareness to apply it appropriately. However, a gap exists in evaluating LLMs’ ability to judge humor suitability within professional settings, especially industry-specific contexts. Figure \ref{fig:examples} illustrates two examples in which LLMs fall short of correctly determining the suitability of industry-specific humor. To fill this gap, we present the first dataset for assessing LLMs' humor appropriateness across various industrial contexts. Our dataset includes 304 annotated humorous statements from industry-specific environments and a brief description of each statement.

\paragraph{Contributions} 1) We introduce the first dataset about humor in industrial settings. 2) We perform a systematic exploration and comparative analysis of LLMs to predict the appropriateness of humorous content. 3) We offer a comprehensive discussion of the identified errors. Our dataset is publicly available at \href{https://anonymous.4open.science/r/PHD-B414/README.md}{repository}.

\section{Related work}
\subsection{Computational Humor Generation}
Computational humor lies at the intersection of Natural Language Processing (NLP) and Humor theory. \citep{ravi2024small} Early work on computational humor generation focused on automated systems that could produce humorous content, but these were often limited to predefined templates and structures. \citep{binsted1994implementedmodelpunningriddles, stock-strapparava-2005-hahacronym, 10.5555/1735862.1735867, 4724484} used fixed templates and rules to generate content like puns and acronyms.

The introduction of LLMs has revolutionized computational humor generation. Recent studies have leveraged LLMs' flexibility and generative capabilities to create diverse, high-quality resources of humorous content. \citep{chen-etal-2024-u} generated a 2.1K dataset of puns in Chinese using LLMs, \citep{Zhong_2024_CVPR} introduced a multimodal resource of humor based on the Oogiri game, and \citep{tikhonov2024humormechanicsadvancinghumor} explored the generation of one-liner jokes using a multi-step reasoning process powered by LLMs. These works highlight LLMs' potential to generate humor at scale, overcoming the limitations of template-based approaches. However, context-specific humor resources are still needed, as the appropriateness of humor depends on the environment.\citep{zhou-etal-2023-cobra} One such environment is the industrial setting, where humor must be used carefully, but no existing resource addresses this need.

\subsection{Computational Humor Detection and Evaluation}
Alongside the advancements in humor generation, researchers have also explored the use of computational techniques for humor detection, classification, and evaluation. Early studies relied on classic machine learning models like Naive Bayes and Support Vector Machines to classify or identify humor. \citep{mihalcea-strapparava-2005-making, yang-etal-2015-humor, 10.1145/2661829.2661997}

The rise of LLMs has raised concerns about proper use and issues like bias, impacting computational humor detection and evaluation. \citep{wu2024humour} uses LLMs for humor classification, \citep{tikhonov2024humor} evaluates Vision-language models on humor, \citep{chen-etal-2024-u} evaluates LLMs on Chinese Pun humor, and \citep{ravi2024small} uses LLMs as evaluator for their Humor Distillation task.

Despite the progress in computational humor research evaluation, LLMs' reasoning for specific environments and settings, such as professional, industry-specific humor, has remained unexplored. This paper focuses on developing resources and techniques to answer this growing need.
\section{Dataset}

The proposed dataset contains humor sentences across various industrial contexts, humor types, and appropriateness. This section details the dataset's construction and description.
\subsection{Dataset framework}
In this subsection, details about each field in the resource is presented.
\paragraph{Sentence} contains LLM-generated humorous sentences tailored for industrial or professional contexts, highlighting common workplace scenarios that employees may find amusing or relatable.

\paragraph{Appropriateness} indicates the sentence's suitability for the given context with for levels. ``Offensive'' contains humor that targets a specific nation, culture, or group through biased content, or is overly harsh in tone; ``Mildly Inappropriate'' is typically humor that shows mild disrespect toward a particular role or department within the company; ``Neutral'' reflects factual observations about the company, avoids targeting specific individuals or departments, and remains free of bias; ``Wholesome'' is similar to Neutral but often carries a lighthearted tone, highlighting positive aspects of the company humorously.


\paragraph{Industry-Specific Context} identifies the industry or setting relevant to the humor, like “Marketing” or “Project Management”.

\paragraph{Humor Type} classifies the humor type, enabling analysis of different humor methods and their appropriateness across contexts. Detailed descriptions of each type are in Appendix \ref{sec:gen_appendix}.

\begin{table}[!ht]
  \centering
  \begin{tabular}{lcc|c}
    \hline
    \textbf{Model} & \textbf{GPT} & \textbf{Claude} & \textbf{Sum} \\
    \hline
    Offensive                   & 36 & 21 & 57 \\
    Mildly inappropriate        & 49 & 27 & 76 \\
    Neutral                     & 36 & 48 & 84 \\
    Wholesome                   & 27 & 60 & 87 \\
    \hline
    \textbf{Sum}                & 148 & 156 & 304 \\
  \end{tabular}
  \caption{Sample counts by the model used and level of appropriateness.}
  \label{tab:model_counts}
\end{table}

\paragraph{Short Explanation} offers a quick interpretation, outlining the joke or message.

\begin{table*}[!t]
\centering
\begin{tabular}{lccccc|ccccc}
\hline
\multicolumn{1}{c}{} & \multicolumn{5}{c}{\textbf{Generated by Claude3.5 Sonnet}} & \multicolumn{5}{c}{\textbf{Generated by GPT-4o}} \\
\hline
\textbf{Model} & \textbf{O} & \textbf{M} & \textbf{N} & \textbf{W} & \textbf{W. Avg.} & \textbf{O} & \textbf{M} & \textbf{N} & \textbf{W} & \textbf{W. Avg.} \\
\hline
GPT-4o & \textbf{0.50} & 0.50 & 0.34 & 0.88 & \textbf{0.60} & - & - & - & - & - \\
Claude3.5 Sonnet & - & - & - & - & - & \textbf{0.74} & \textbf{0.60} &\textbf{ 0.86} & 0.77 & \textbf{0.73} \\
Gemini 1.5 Flash & 0.17 & 0.41 & 0.00 & \textbf{0.99} & 0.48 & 0.11 & \textbf{0.60} & 0.00 & \textbf{1.0} & 0.41 \\
Llama-3.2-1B-Instruct & 0.44 & 0.46 & \textbf{0.77} & 0.52 & 0.58 & 0.20 & 0.52 & 0.80 & 0.58 & 0.52  \\
Qwen2.5-72B-Instruct & 0.17 & \textbf{0.54} & 0.15 & 0.97 & 0.54 & 0.00 & 0.47 & 0.05 & 0.96 & 0.34 \\
\hline
Average & 0.32 & 0.48 & 0.31 & \textbf{0.84} & \textbf{0.55} & 0.26 & 0.55 & 0.43 & \textbf{0.83} & 0.50 \\
\end{tabular}
\caption{F1-score of 5 LLMs on the two subsets of the resource. O, M, N, and W are Offensive, Mildly Inappropriate, Neutral, and Wholesome respectively. W. Avg. stands for Weighted average.}
\label{tab:experiment_results}
\end{table*}

\begin{table}[!ht]
  \centering
  \begin{tabular}{lc}
    \hline
    \textbf{Humor type} & \textbf{Count} \\
    \hline
    Irony     & 59 \\
    Hyperbole     & 26 \\
    Self-deprecation     & 16 \\
    Metaphorical Humor     & 43 \\
    Situational Humor     & 39 \\
    Positive Humor     & 38 \\
    Cultural Reference Humor     & 83 \\
    \hline
  \end{tabular}
  \caption{Sample counts by humor type.}
  \label{tab:humor_type}
\end{table}

\subsection{Humorous content generation}
Sentences along with their associated features were initially generated through LLM prompting, available in Appendix \ref{sec:gen_appendix}. The prompt consists of instances of industrial settings along with a description of each feature. Additionally, several rules were established to ensure distinctiveness, diversity, and conciseness, and to prevent redundancy. Also, each prompt consists of a sequence of following prompts, which emphasize different aspects of the resource, including cultural, industrial, and appropriateness diversity.
After the careful process of designing the prompts, Claude 3.5 Sonnet \footnote{\href{https://www.anthropic.com/}{https://www.anthropic.com/}} and ChatGPT-4o \footnote{\href{https://chatgpt.com/}{https://chatgpt.com/}} were used to generate the initial set of samples. This procedure resulted in 340 samples, 170 from each model. Two experienced annotators then reviewed the quality of generated samples and removed irrelevant, repetitive, or nonsensical ones. After this step, the size was reduced to 304.

\subsection{Annotation}
During sample generation, both LLMs were prompted to not only create but also label examples to ensure diversity by type, level, and features, though these initial labels were later discarded. Two male annotators independently assessed each sample for appropriateness, required knowledge, industry context, and humor type. For consistency, they standardized labels when samples described the same context differently. In cases of disagreement between annotators, a third male reviewer was consulted to resolve conflicts.

\subsection{Description}
The resource includes 304 samples, with 138 different industry-specific contexts. The overall distribution of appropriateness levels, humor types, and counts of each model is presented in Tables \ref{tab:model_counts} and \ref{tab:humor_type}.

\section{Experiments}
\textbf{Data} Our experiments center around the professional, industry-specific dataset introduced in this work. For each experiment, we craft a prompt that includes a general dataset description, explanations of appropriateness levels, and details of the appropriateness classification task. This prompt is paired with a humorous sentence from the dataset and then given to the model. A complete version of the prompt is available in \ref{sec:ex_appendix}.

\textbf{Model} In addition to Claude3.5 Sonnet and ChatGPT-4o, we evaluated three additional models, two of which are openly accessible: Llama-3.2-1B-Instruct \citep{meta2024llama}, Qwen2.5-72B-Instruct \citep{yang2024qwen2}, and Gemini 1.5 Flash\footnote{\href{https://gemini.google.com/}{https://gemini.google.com/}}. Since the two first models were used in the initial data generation step, each was only tested on the subset they were not involved in generating. We set the temperature to zero in all experiments to ensure deterministic model responses. All responses were collected in November 2024.

\section{Results}

\begin{table}[!ht]
  \centering
  \begin{tabular}{lcccc|c}
    \hline
    \textbf{Type} & \textbf{O} & \textbf{M} & \textbf{N} & \textbf{W} & \textbf{W A} \\
    \hline
    Irony                   & \textbf{0.70} & 0.51 & 0.80 & 0.25 & \textbf{0.60} \\
    Hyperbole               & 0.33 & 0.33 & \textbf{0.82} & \textbf{0.80} & \textbf{0.60} \\
    Self-dep.               & 0.00 & 0.40 & 0.57 & 0.00 & 0.38 \\
    Metaphor                & 0.33 & \textbf{0.60} & 0.75 & 0.33 & 0.56 \\
    Situational             & 0.00 & 0.44 & 0.84 & 0.46 & 0.56 \\
    Positive                & - & - & 0.80 & 0.54 & 0.56 \\
    Cultural Ref.           & 0.07 & 0.53 & 0.75 & 0.73 & 0.41 \\
    \hline
  \end{tabular}
  \caption{F1-score of Llama-3.2-1B-Instruct across different types of humor and levels of appropriateness. O, M, N, and W are Offensive, Mildly Inappropriate, Neutral, and Wholesome respectively. W A stands for Weighted average.}
  \label{tab:type_level_rel}
\end{table}

The results of our analysis in Table \ref{tab:experiment_results} reveal that both Claude3.5 Sonnet and ChatGPT-4o can determine the appropriateness level of the subset that they were not involved in the generation better than the rest of the models. Apart from the two mentioned models, Llama-3.2-1B-Instruct performs better than Qwen and Gemini.
Furthermore, the average f1-score of models on the subset generated by GPT-4o is lower than that of the Claude subset, suggesting it contains more nuanced, context-specific humor that demands a deeper understanding of professional dynamics and workplace culture.
The offensive category has low F1-scores in both subsets, indicating that models may overlook cultural or national references, revealing potential biases. For instance,\textit{ “In our Italian office, deadlines are merely suggestions – like the speed limit on a Sicilian road”} is offensive toward Italian culture, yet Qwen and Llama classify it as Neutral. The Neutral category also has low F1-scores due to its broad, flexible nature. In contrast, Wholesome and Mildly Inappropriate achieve better scores, especially Wholesome, suggesting these categories are easier for LLMs to detect due to their distinctive boundaries.
We choose Llama-3.2-1B-Instruct as the best model since among the models tested on both subsets to analyze the relationship between humor type and models' ability to predict appropriateness levels.

Table \ref{tab:type_level_rel} presents the f1-score of the best model across all humor types for each appropriateness level. The natural relationship between each humor type and each level of appropriateness can explain the scores. Irony achieves the highest F1-score in the Offensive category, likely due to its subtle contradictions or contrasts, which can sometimes be interpreted as offensive, especially when highlighting flaws. In the Mildly Inappropriate category, metaphorical humor scores the highest, as metaphors often introduce suggestive comparisons that come close to inappropriateness without being overtly offensive. Hyperbole achieves top scores in both Neutral and Wholesome categories, as its exaggerated statements are humorous yet boundary-respecting, making it well-suited for neutral and wholesome contexts.

\section{Conclusion}
We present the first resource focused on humor within industrial contexts, encompassing a diverse range of humor types and levels of appropriateness. Each sample is tied to a specific context, which shapes its unique humorous character. This work is timely, as humor studies are gaining momentum due to the potential applications of humor in automated tasks, such as email writing. Our goal is to promote human-aligned humor, enabling AI assistants to communicate in a more relatable, human-like manner, by focusing on the intersection of NLP and humor theory.
\section{Limitations}
The present work is not without limitations. Mainly, there are three important limitations associated with the work. 

Firstly, the dataset size is limited. While we made efforts to create a valuable resource with a diverse range of contexts, we took care to minimize redundancy by avoiding the creation of samples with similar contexts, meanings, and labels. Nevertheless, the dataset currently contains 304 samples, which presents an opportunity for future expansion.

Additionally, our classification of humor types has some limitations and could be expanded to capture more nuances. For instance, we grouped samples that could be classified as sarcasm under the broader category of irony. This was done to maintain a minimal set of types, as some categories, such as sarcasm, had limited samples in the initial dataset.

Lastly, as both annotators are male we removed all samples containing gender references to avoid bias.

\section{Ethical Considerations}
To streamline dataset compilation, we used LLMs to generate potential humorous examples. While this significantly enhanced efficiency, it may also introduce biases. The humor generated by LLMs could differ from what human experts might select, potentially reflecting biases inherent in the model or prompt design.

The annotation process was carried out by two experienced annotators with industry experience. However, since both annotators are male, there is potential for bias, despite our efforts to eliminate gender references in order to minimize this effect.

\bibliography{custom}

\begin{thebibliography}{39}
\providecommand{\natexlab}[1]{#1}

\bibitem[{Banas et~al.(2011)Banas, Dunbar, Rodriguez, and Liu}]{banas2011review}
John~A Banas, Norah Dunbar, Dariela Rodriguez, and Shr-Jie Liu. 2011.
\newblock A review of humor in educational settings: Four decades of research.
\newblock \emph{Communication Education}, 60(1):115--144.

\bibitem[{Bardon(2005)}]{bardon2005philosophy}
Adrian Bardon. 2005.
\newblock The philosophy of humor.
\newblock \emph{Comedy: A geographic and historical guide}, 2:462--476.

\bibitem[{Berk(2015)}]{berk2015greatest}
Ronald~A Berk. 2015.
\newblock The greatest veneration: Humor as a coping strategy for the challenges of aging.
\newblock \emph{Social Work in Mental Health}, 13(1):30--47.

\bibitem[{Binsted and Ritchie(1994)}]{binsted1994implementedmodelpunningriddles}
Kim Binsted and Graeme Ritchie. 1994.
\newblock \href {https://arxiv.org/abs/cmp-lg/9406022} {An implemented model of punning riddles}.
\newblock \emph{Preprint}, arXiv:cmp-lg/9406022.

\bibitem[{Chen et~al.(2024)Chen, Yang, Hu, Chen, Lan, Cai, Zhuang, Lin, Lu, and Zhou}]{chen-etal-2024-u}
Yang Chen, Chong Yang, Tu~Hu, Xinhao Chen, Man Lan, Li~Cai, Xinlin Zhuang, Xuan Lin, Xin Lu, and Aimin Zhou. 2024.
\newblock \href {https://doi.org/10.18653/v1/2024.findings-acl.51} {Are {U} a joke master? pun generation via multi-stage curriculum learning towards a humor {LLM}}.
\newblock In \emph{Findings of the Association for Computational Linguistics: ACL 2024}, pages 878--890, Bangkok, Thailand. Association for Computational Linguistics.

\bibitem[{Coser(1959)}]{coser1959some}
Rose~Laub Coser. 1959.
\newblock Some social functions of laughter: A study of humor in a hospital setting.
\newblock \emph{Human relations}, 12(2):171--182.

\bibitem[{Dybala et~al.(2010)Dybala, Ptaszynski, Maciejewski, Takahashi, Rzepka, and Araki}]{10.5555/1735862.1735867}
Pawel Dybala, Michal Ptaszynski, Jacek Maciejewski, Mizuki Takahashi, Rafal Rzepka, and Kenji Araki. 2010.
\newblock Multiagent system for joke generation: Humor and emotions combined in human-agent conversation.
\newblock \emph{J. Ambient Intell. Smart Environ.}, 2(1):31–48.

\bibitem[{Etxaniz et~al.(2024)Etxaniz, Azkune, Soroa, de~Lacalle, and Artetxe}]{etxaniz2024bertaqa}
Julen Etxaniz, Gorka Azkune, Aitor Soroa, Oier~Lopez de~Lacalle, and Mikel Artetxe. 2024.
\newblock \href {https://openreview.net/forum?id=QocjHRR31U} {Berta{QA}: How much do language models know about local culture?}
\newblock In \emph{The Thirty-eight Conference on Neural Information Processing Systems Datasets and Benchmarks Track}.

\bibitem[{Gryka et~al.(2024)Gryka, Grado{\'n}, Koz{\l}owski, Kuty{\l}a, and Janicki}]{gryka2024detection}
Pawe{\l} Gryka, Kacper Grado{\'n}, Marek Koz{\l}owski, Mi{\l}osz Kuty{\l}a, and Artur Janicki. 2024.
\newblock Detection of ai-generated emails-a case study.
\newblock In \emph{Proceedings of the 19th International Conference on Availability, Reliability and Security}, pages 1--8.

\bibitem[{Hampes(1992)}]{hampes1992relation}
William~P Hampes. 1992.
\newblock Relation between intimacy and humor.
\newblock \emph{Psychological reports}, 71(1):127--130.

\bibitem[{Hong and Ong(2008)}]{4724484}
Bryan~Anthony Hong and Ethel Ong. 2008.
\newblock \href {https://doi.org/10.1109/ISUC.2008.28} {Generating punning riddles from examples}.
\newblock In \emph{2008 Second International Symposium on Universal Communication}, pages 347--352.

\bibitem[{Idrees et~al.(2020)Idrees, Batool, and Kausar}]{idrees2020styles}
Ayesha Idrees, Saira Batool, and Rukhsana Kausar. 2020.
\newblock Styles of humor and interpersonal relationships in university students.
\newblock \emph{FWU Journal of Social Sciences}, 14(4):57--67.

\bibitem[{Jiang et~al.(2019)Jiang, Li, and Hou}]{jiang2019cultural}
Tonglin Jiang, Hao Li, and Yubo Hou. 2019.
\newblock Cultural differences in humor perception, usage, and implications.
\newblock \emph{Frontiers in psychology}, 10:123.

\bibitem[{Jovic and Mnasri(2024)}]{jovic2024evaluating}
Marina Jovic and Salaheddine Mnasri. 2024.
\newblock Evaluating ai-generated emails: A comparative efficiency analysis.
\newblock \emph{World Journal of English Language}, 14(2).

\bibitem[{Lu et~al.(2024)Lu, Mysore, Safavi, Neville, Yang, and Wan}]{lu2024corporate}
Zhuoran Lu, Sheshera Mysore, Tara Safavi, Jennifer Neville, Longqi Yang, and Mengting Wan. 2024.
\newblock Corporate communication companion (ccc): An llm-empowered writing assistant for workplace social media.
\newblock \emph{arXiv preprint arXiv:2405.04656}.

\bibitem[{Lynch(2002)}]{lynch2002humorous}
Owen~H Lynch. 2002.
\newblock Humorous communication: Finding a place for humor in communication research.
\newblock \emph{Communication theory}, 12(4):423--445.

\bibitem[{Mesmer-Magnus et~al.(2012)Mesmer-Magnus, Glew, and Viswesvaran}]{mesmer2012meta}
Jessica Mesmer-Magnus, David~J Glew, and Chockalingam Viswesvaran. 2012.
\newblock A meta-analysis of positive humor in the workplace.
\newblock \emph{Journal of Managerial Psychology}, 27(2):155--190.

\bibitem[{Meta et~al.(2024)Meta, Jauhri, Pandey, Kadian, Al-Dahle, Letman, Mathur, Schelten, Yang, Fan et~al.}]{meta2024llama}
AI~Meta, Abhinav Jauhri, Abhinav Pandey, Abhishek Kadian, Ahmad Al-Dahle, Aiesha Letman, Akhil Mathur, Alan Schelten, Amy Yang, Angela Fan, et~al. 2024.
\newblock The llama 3 herd of models.
\newblock \emph{arXiv preprint arXiv:2407.21783}, 2.

\bibitem[{Meyer(2000)}]{meyer2000humor}
John~C Meyer. 2000.
\newblock Humor as a double-edged sword: Four functions of humor in communication.
\newblock \emph{Communication theory}, 10(3):310--331.

\bibitem[{Meyer(2015)}]{meyer2015understanding}
John~C Meyer. 2015.
\newblock \emph{Understanding humor through communication: Why be funny, anyway?}
\newblock Lexington Books.

\bibitem[{Mihalcea and Strapparava(2005)}]{mihalcea-strapparava-2005-making}
Rada Mihalcea and Carlo Strapparava. 2005.
\newblock \href {https://aclanthology.org/H05-1067} {Making computers laugh: Investigations in automatic humor recognition}.
\newblock In \emph{Proceedings of Human Language Technology Conference and Conference on Empirical Methods in Natural Language Processing}, pages 531--538, Vancouver, British Columbia, Canada. Association for Computational Linguistics.

\bibitem[{Morreall(1983)}]{morreall1983humor}
John Morreall. 1983.
\newblock Humor and emotion.
\newblock \emph{American Philosophical Quarterly}, 20(3):297--304.

\bibitem[{Park et~al.(2023)Park, O'Brien, Cai, Morris, Liang, and Bernstein}]{park2023generative}
Joon~Sung Park, Joseph O'Brien, Carrie~Jun Cai, Meredith~Ringel Morris, Percy Liang, and Michael~S Bernstein. 2023.
\newblock Generative agents: Interactive simulacra of human behavior.
\newblock In \emph{Proceedings of the 36th annual acm symposium on user interface software and technology}, pages 1--22.

\bibitem[{Ravi et~al.(2024)Ravi, Huber, Shrivastava, Sagar, Aly, Shwartz, and Einolghozati}]{ravi2024small}
Sahithya Ravi, Patrick Huber, Akshat Shrivastava, Aditya Sagar, Ahmed Aly, Vered Shwartz, and Arash Einolghozati. 2024.
\newblock Small but funny: A feedback-driven approach to humor distillation.
\newblock \emph{arXiv preprint arXiv:2402.18113}.

\bibitem[{Scherbakov et~al.(2024)Scherbakov, Hubig, Jansari, Bakumenko, and Lenert}]{scherbakov2024emergence}
Dmitry Scherbakov, Nina Hubig, Vinita Jansari, Alexander Bakumenko, and Leslie~A Lenert. 2024.
\newblock The emergence of large language models (llm) as a tool in literature reviews: an llm automated systematic review.
\newblock \emph{arXiv preprint arXiv:2409.04600}.

\bibitem[{Shen et~al.(2024)Shen, Li, Wang, and Qu}]{shen2024data}
Leixian Shen, Haotian Li, Yun Wang, and Huamin Qu. 2024.
\newblock From data to story: Towards automatic animated data video creation with llm-based multi-agent systems.
\newblock \emph{arXiv preprint arXiv:2408.03876}.

\bibitem[{Stock and Strapparava(2005)}]{stock-strapparava-2005-hahacronym}
Oliviero Stock and Carlo Strapparava. 2005.
\newblock \href {https://doi.org/10.3115/1225753.1225782} {{HAHA}cronym: A computational humor system}.
\newblock In \emph{Proceedings of the {ACL} Interactive Poster and Demonstration Sessions}, pages 113--116, Ann Arbor, Michigan. Association for Computational Linguistics.

\bibitem[{Tikhonov and Shtykovskiy(2024{\natexlab{a}})}]{tikhonov2024humormechanicsadvancinghumor}
Alexey Tikhonov and Pavel Shtykovskiy. 2024{\natexlab{a}}.
\newblock \href {https://arxiv.org/abs/2405.07280} {Humor mechanics: Advancing humor generation with multistep reasoning}.
\newblock \emph{Preprint}, arXiv:2405.07280.

\bibitem[{Tikhonov and Shtykovskiy(2024{\natexlab{b}})}]{tikhonov2024humor}
Alexey Tikhonov and Pavel Shtykovskiy. 2024{\natexlab{b}}.
\newblock Humor mechanics: Advancing humor generation with multistep reasoning.
\newblock \emph{arXiv preprint arXiv:2405.07280}.

\bibitem[{Vizmuller(1980)}]{vizmuller1980psychological}
Jana Vizmuller. 1980.
\newblock Psychological reasons for using humor in a pedagogical setting.
\newblock \emph{Canadian modern language review}, 36(2):266--271.

\bibitem[{Wang et~al.(2024)Wang, Ma, Feng, Zhang, Yang, Zhang, Chen, Tang, Chen, Lin et~al.}]{wang2024survey}
Lei Wang, Chen Ma, Xueyang Feng, Zeyu Zhang, Hao Yang, Jingsen Zhang, Zhiyuan Chen, Jiakai Tang, Xu~Chen, Yankai Lin, et~al. 2024.
\newblock A survey on large language model based autonomous agents.
\newblock \emph{Frontiers of Computer Science}, 18(6):186345.

\bibitem[{Wang(2024)}]{wang2024investigating}
Shan Wang. 2024.
\newblock Investigating the potential of large language models for automated writing scoring.
\newblock In \emph{2024 5th International Conference on Education, Knowledge and Information Management (ICEKIM 2024)}, pages 1091--1098. Atlantis Press.

\bibitem[{Wu et~al.(2024)Wu, Huang, and Lau}]{wu2024humour}
Shih-Hung Wu, Yu-Feng Huang, and Tsz-Yeung Lau. 2024.
\newblock Humour classification by fine-tuning llms: Cyut at clef 2024 joker lab subtask humour classification according to genre and technique.
\newblock In \emph{Working Notes of the Conference and Labs of the Evaluation Forum (CLEF 2024). CEUR Workshop Proceedings}, pages 1933--1947.

\bibitem[{Xiao et~al.(2024)Xiao, Ma, Xu, Zhang, Wang, and Fu}]{xiao2024automation}
Changrong Xiao, Wenxing Ma, Sean~Xin Xu, Kunpeng Zhang, Yufang Wang, and Qi~Fu. 2024.
\newblock From automation to augmentation: Large language models elevating essay scoring landscape.
\newblock \emph{arXiv preprint arXiv:2401.06431}.

\bibitem[{Yang et~al.(2024)Yang, Yang, Hui, Zheng, Yu, Zhou, Li, Li, Liu, Huang et~al.}]{yang2024qwen2}
An~Yang, Baosong Yang, Binyuan Hui, Bo~Zheng, Bowen Yu, Chang Zhou, Chengpeng Li, Chengyuan Li, Dayiheng Liu, Fei Huang, et~al. 2024.
\newblock Qwen2 technical report.
\newblock \emph{arXiv preprint arXiv:2407.10671}.

\bibitem[{Yang et~al.(2015)Yang, Lavie, Dyer, and Hovy}]{yang-etal-2015-humor}
Diyi Yang, Alon Lavie, Chris Dyer, and Eduard Hovy. 2015.
\newblock \href {https://doi.org/10.18653/v1/D15-1284} {Humor recognition and humor anchor extraction}.
\newblock In \emph{Proceedings of the 2015 Conference on Empirical Methods in Natural Language Processing}, pages 2367--2376, Lisbon, Portugal. Association for Computational Linguistics.

\bibitem[{Zhang and Liu(2014)}]{10.1145/2661829.2661997}
Renxian Zhang and Naishi Liu. 2014.
\newblock \href {https://doi.org/10.1145/2661829.2661997} {Recognizing humor on twitter}.
\newblock In \emph{Proceedings of the 23rd ACM International Conference on Conference on Information and Knowledge Management}, CIKM '14, page 889–898, New York, NY, USA. Association for Computing Machinery.

\bibitem[{Zhong et~al.(2024)Zhong, Huang, Gao, Wen, Lin, Zitnik, and Zhou}]{Zhong_2024_CVPR}
Shanshan Zhong, Zhongzhan Huang, Shanghua Gao, Wushao Wen, Liang Lin, Marinka Zitnik, and Pan Zhou. 2024.
\newblock Let's think outside the box: Exploring leap-of-thought in large language models with creative humor generation.
\newblock In \emph{Proceedings of the IEEE/CVF Conference on Computer Vision and Pattern Recognition (CVPR)}, pages 13246--13257.

\bibitem[{Zhou et~al.(2023)Zhou, Zhu, Yerukola, Davidson, Hwang, Swayamdipta, and Sap}]{zhou-etal-2023-cobra}
Xuhui Zhou, Hao Zhu, Akhila Yerukola, Thomas Davidson, Jena~D. Hwang, Swabha Swayamdipta, and Maarten Sap. 2023.
\newblock \href {https://doi.org/10.18653/v1/2023.findings-acl.392} {{COBRA} frames: Contextual reasoning about effects and harms of offensive statements}.
\newblock In \emph{Findings of the Association for Computational Linguistics: ACL 2023}, pages 6294--6315, Toronto, Canada. Association for Computational Linguistics.

\end{thebibliography}

\appendix

\section{Generation of the data}
\label{sec:gen_appendix}
In this appendix, we provide a detailed overview of the process for generating the initial samples. 
First, various types of humor type in the dataset are explained below:
\begin{itemize}
    \item \textbf{Irony} conveys humor through contrast or subtle contradictions that highlight a disparity between expectations and reality.
    \item \textbf{Hyperbole} uses exaggeration to humorously emphasize a point, often to highlight absurdity or a commonly shared frustration.
    \item \textbf{Self-deprecation humor} employs humility, allowing the speaker to make light of their own flaws or challenges in a relatable way.
    \item \textbf{Metaphorical Humor} uses creative comparisons to convey ideas, often making complex or dry topics more engaging and accessible.
    \item \textbf{Situational Humor} draws humor from specific workplace scenarios or events, creating relatability through common professional experiences.
    \item \textbf{Positive Humor} emphasizes lighthearted, uplifting jokes that foster a positive atmosphere without targeting anyone.
    \item \textbf{Cultural Reference Humor} incorporates widely known cultural elements, allowing the audience to connect over shared knowledge or experiences.
\end{itemize}

Moreover, we provide the prompt we used to generate the initial dataset. Below, the generation prompt is available:

\texttt{I am developing a dataset of Professional industries-specific Humor. 
In the following paragraphs, I will provide some general contexts in which professional humor might appear: Corporate, communications, Marketing, campaigns, Professional presentations, and Email threads.}

\texttt{Also, here are the columns in my dataset:
\begin{itemize}
    \item \textbf{Sentence}: The humorous sentence reflects humor within an industrial or professional context. The sentence aims to highlight a common workplace or industry-specific content in a way that employees can find amusing or relatable.
    \item \textbf{Appropriateness level}: Indicates the sentence's suitability for the given context with four levels. ``Offensive'' contains humor that targets a specific nation, culture, or group through biased content, or is overly harsh in tone; ``Mildly Inappropriate'' is typically humor that shows mild disrespect toward a particular role or department within the company; ``Neutral'' reflects factual observations about the company, avoids targeting specific individuals or departments, and remains free of bias; ``Wholesome'' is similar to Neutral but often carries a lighthearted tone, highlighting positive aspects of the company humorously.
    \item \textbf{Industry-specific context}: Identifies the specific industry or professional setting relevant to the humor, such as “Marketing” or “Project Management”.
    \item \textbf{Humor type}: categorizes the type of humor, with various categories. ``Irony'' conveys humor through contrast or subtle contradictions that highlight a disparity between expectations and reality; ``Hyperbole'' uses exaggeration to humorously emphasize a point, often to highlight absurdity or a commonly shared frustration; ``Self-deprecation'' employs humility, allowing the speaker to make light of their own flaws or challenges in a relatable way; ``Metaphorical Humor'' uses creative comparisons to convey ideas, often making complex or dry topics more engaging and accessible; ``Situational Humor'' draws humor from specific workplace scenarios or events, creating relatability through common professional experiences; ``Positive Humor'' emphasizes lighthearted, uplifting jokes that foster a positive atmosphere without targeting anyone; ``Cultural Reference Humor'' incorporates widely known cultural elements, allowing the audience to connect over shared knowledge or experiences.
    \item \textbf{Short explanation}: provides a brief interpretation of the humor, outlining the underlying joke or message. This explanation aids in annotating the dataset, helping reviewers understand the humor’s intent and contextual relevance, which is essential for quality control and evaluation.
\end{itemize}
}

\texttt{There are some generation rules: 
\begin{itemize}
    \item Generate as distinct samples as possible.
    \item Focus on different industries and create diversity in the dataset.
    \item Use the same words for similar classes, types, etc, when possible.
    \item Be concise.
    \item You must return in a CSV-like format, where the separator is "|".
\end{itemize}
Now, generate 10 samples based on the given description. 
}

Additionally, to increase the diversity of generated samples, we occasionally appended one of several specific sentences to the end of the prompt.

\texttt{
\begin{itemize}
    \item Focus on different cultures and countries. 
    \item Focus on different industries and companies.
    \item Focus on different levels of Appropriateness.
    \item Focus on different types of humor.
    \item Focus on those that require specific background knowledge.
    \item Focus on those with the potential of being offensive.
    \item Focus on those with the potential of being Mildly Inappropriate.
    \item Focus on those with the potential of being Neutral.
    \item Focus on those with the potential to be Wholesome.
\end{itemize}
}

\section{Evaluation Prompt}
\label{sec:ex_appendix}
This appendix provides additional details on our experiments, focusing on the prompt used, as outlined below.
\texttt{We have a dataset of humorous sentences in professional industrial settings. We want you to classify the given sentences based on their appropriateness level. There are four levels of appropriateness:
\begin{itemize}
    \item ``Offensive'' contains humor that targets a specific nation, culture, or group through biased content, or is overly harsh in tone.
    \item ``Mildly Inappropriate'' is typically humor that shows mild disrespect toward a particular role or department within the company.
    \item ``Neutral'' reflects factual observations about the company, avoids targeting specific individuals or departments, and remains free of bias.
    \item ``Wholesome'' is similar to Neutral but often carries a lighthearted tone, highlighting positive aspects of the company humorously.
\end{itemize}
Classify each sentence according to the provided description. Only state the level without further explanation.}
\end{document}